\documentclass[conference]{IEEEtran}
\ifCLASSINFOpdf
\else
\fi

\usepackage{graphicx}
\usepackage{subcaption}
\usepackage{array} 
\usepackage{colortbl} 
\usepackage{hhline} 
\usepackage{booktabs}
\usepackage{amsmath}
\usepackage{afterpage}
\usepackage [utf8]{inputenc}
\usepackage [T1]{fontenc}
\usepackage {geometry}
\usepackage {atbegshi}
\usepackage {lipsum}

\makeatletter
\gdef\SecondPageGeometry 
{\ifnum 0=\currentgrouplevel 
\global\let\SecondPageGeometry =\relax 
\expandafter \@gobble 
\newgeometry {top=19.1mm,left=19.1mm,
right= 19.1mm,bottom =20mm}%
\else \aftergroup \SecondPageGeometry \fi 
}

\AtBeginShipout {\AtBeginShipoutUpperLeft 
{\SecondPageGeometry }}

\begin{document}

\newcolumntype{I}{!{\vrule width 1.5pt}}

%
\newgeometry {top=25.4mm,left=19.1mm, right= 19.1mm,bottom =20mm}

\title{AV-Occupant Perceived Risk Model for Cut-In Scenarios with Empirical Evaluation}



\author{\IEEEauthorblockN{Sarah Barendswaard}
\IEEEauthorblockA{Siemens Digital Industries Software\\
Email: sarah.barendswaard@siemens.com}
\and
\IEEEauthorblockN{Tong Duy Son}
\IEEEauthorblockA{Siemens Digital Industries Software\\
Email: son.tong@siemens.com}}


%


\maketitle

\begin{abstract}
Advancements in autonomous vehicle (AV) technologies necessitate precise estimation of perceived risk to enhance user comfort, acceptance and trust. This paper introduces a novel AV-Occupant Risk (AVOR) model designed for perceived risk estimation during AV cut-in scenarios. An empirical study is conducted with 18 participants with realistic cut-in scenarios. Two factors were investigated: scenario risk and scene population. 76\% of subjective risk responses indicate an increase in perceived risk at cut-in initiation. The existing perceived risk model did not capture this critical phenomenon. Our AVOR model demonstrated a significant improvement in estimating perceived risk during the early stages of cut-ins, especially for the high-risk scenario, enhancing modelling accuracy by up to 54\%. The concept of the AVOR model can quantify perceived risk in other diverse driving contexts characterized by dynamic uncertainties, enhancing the reliability and human-centred focus of AV systems.
\end{abstract}

%
\IEEEpeerreviewmaketitle
\section{Introduction}
The emergence of autonomous vehicles (AV) promises transformative changes in transportation, offering to improve safety, comfort and reduce congestion \cite{ERTRAC2019}. While improvements in safety and traffic flow can be reliably predicted, assessing comfort remains challenging, requiring prototype testing  due to the lack of quantitative models for its evaluation \cite{Baekgyu2017,SOTIF}. Comfort evaluation is intricate, given its subjective nature, complexity, and individual variability. Yet, comfort has become a crucial criterion in determining the quality of intelligent transportation systems \cite{GUO2021240}, with ISO 2631 providing a method to categorise passenger comfort \cite{ISO2631}. 


Modelling AV occupant comfort presents a complex, multidimensional challenge, primarily encompassing motion comfort and perceived risk. Motion comfort is influenced by 1) ride smoothness—affected by high-frequency motion profiles  and vehicle handling qualities \cite{Mohajer2020,deWinkel2022}—and 2) motion sickness, which arises from discrepancies between visual and vestibular sensory inputs \cite{Irmak2022}. This paper, however, focuses on perceived risk, anchored in the psychological domain of trust and confidence in the AV's capabilities \cite{PRASETIO2023}.

Perceived risk denotes the subjective assessment of potential hazards and their severity. It is a critical factor influencing user willingness to adopt AVs, as high perceived risk can lead to distrust, prompting manual interventions or outright rejection \cite{KOLEKAR2021, Xu2018, MONTORO2019}. Perceived risk and actual risk are, however, distinct concepts \cite{DelRe2022} \cite{CHARLTON201450}. The former is a subjective interpretation often influenced by individual and situational factors. Whereas the latter is objectively defined by severity (of potential injuries), exposure (probability of occurrence under various operating conditions), and controllability (likelihood of driver control effectiveness) as per ISO 26262 \cite{iso26262}. Note that controllability is less relevant for autonomous vehicles (AVs). Actual risk in AV contexts is quantified using surrogate safety measures such as Time-To-Collision (TTC), Time-Headway (THW), and time-to-closest-encounter, among other threat assessment techniques  \cite{Kondoh2008, Tariq2023,Li2021}. These metrics, while effective in estimating objective risk, do not necessarily align with the human perception of risk.

Existing models quantifying perceived risk have limitations. A regression model proposed in \cite{HE2022} estimates perceived risk and trust, but is limited to a one-dimensional perspective, ignoring risks from vehicles outside the immediate lane. Another model by \cite{MULLAKKALBABU2020102716} calculates a probabilistic energy value for collisions, considering kinetic and potential risks, yet it focuses only on collision probability and omits critical factors like lane width and road type besides being computationally intensive. The Driver's Risk Field (DRF) risk model \cite{KOLEKAR2021} estimates risk using ego vehicle parameters and an environmental cost map, accommodating variable road conditions such as lane width changes. However, it inadequately addresses the kinematics and uncertainties of dynamic obstacles, treating static and dynamic obstacles equally. In fact, the cost or the consequence of a static obstacle could be the same as that of a dynamic cut-in obstacle, which proves to be problematic.

%

This work presents a novel Autonomous Vehicle-Occupant Risk (AVOR) model for real-life cut-in scenarios, aiming to improve perceived risk estimation by overcoming existing model limitations in handling diverse traffic conditions, physical infrastructure, and uncertainties of dynamic objects. Building upon the DRF risk model, our enhanced AVOR model integrates dynamic object uncertainties into the environmental cost map. Specifically, we target cut-in manoeuvres, unaddressed by the DRF yet critical in highway and urban collisions \cite{1994NHTSA}. We empirically validate the AVOR model using real-time subjective risk ratings from user experiments \cite{MORAN2019309} \cite{DEWINTER2023235}, incorporating two cut-in scenarios from the Waymo-Open-Dataset processed through our Real2Sim pipeline for simulation in Simcenter Prescan. The objectives of the study are threefold: (1) develop the AVOR model for cut-in manoeuvres; (2) empirically validate the model against drivers' real-time subjective risk perceptions; and (3) investigate the impact of road population and furniture on subjective risk assessments and the model's accuracy in these estimations.

The subsequent sections include a detailed examination of the cut-in maneuver (Section \ref{sec:anatomy}) and the introduction of the developed AV-Occupant risk estimation model (Section \ref{sec:model}). The experimental setup and involved variables are outlined (Section \ref{sec:experiment}). Findings and their discussion are presented in Sections \ref{sec:results} and \ref{sec:discussion}, respectively. The paper concludes with final remarks in Section \ref{sec:conclusion}.

\section{The Anatomy of a Cut-In}
\label{sec:anatomy}

\begin{figure}[htbp]
\centering
\includegraphics[width=0.3\textwidth]{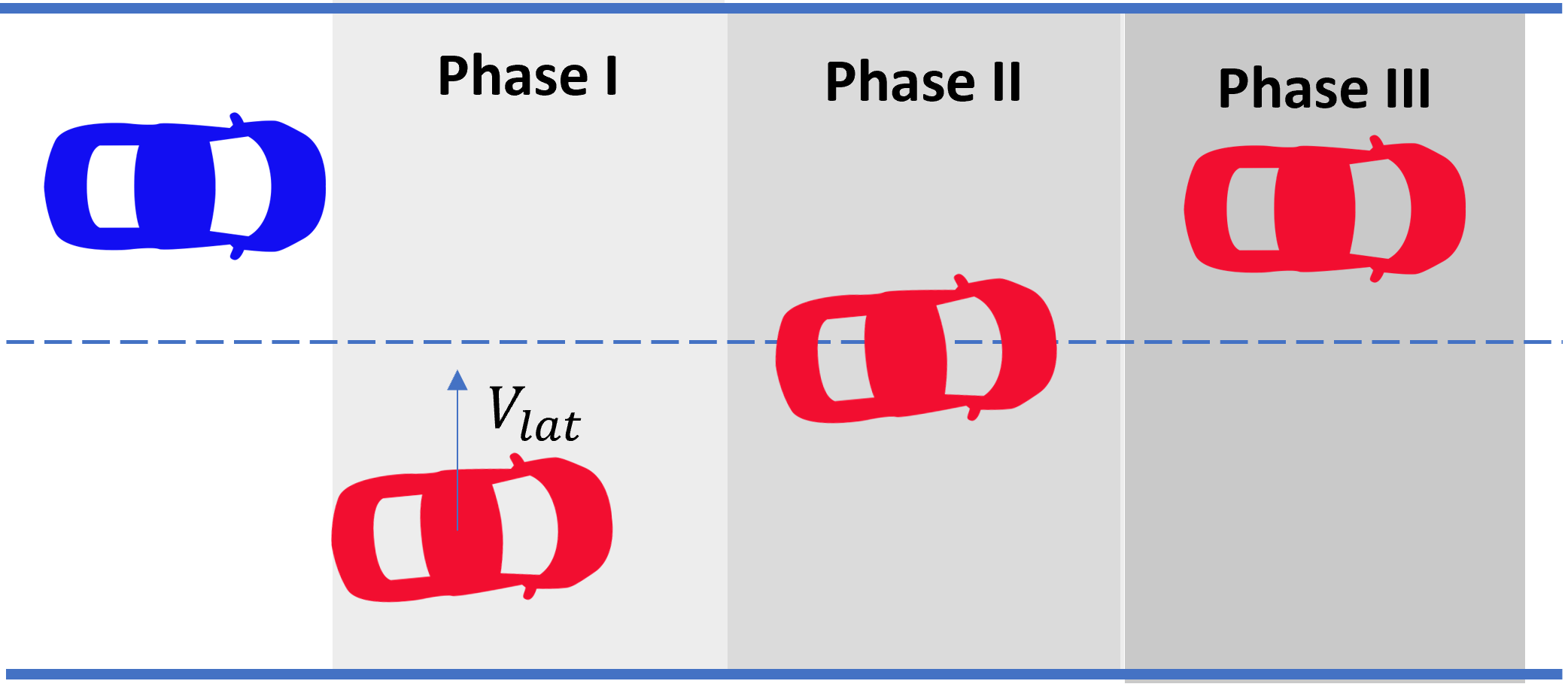}
\caption{Three phases of a cut-in scenario. The ego vehicle is illustrated in blue, and the cut-in vehicle is depicted in red color.}
\label{fig:cut-in_anatomy}
\end{figure}

This article proposes quantifying the cut-in manoeuvre in three distinct phases for more granular analysis, as illustrated in Fig. \ref{fig:cut-in_anatomy}. The scenario data is sectioned into three phases. Phase I is \textit{initiation}: the obstacle vehicle (red) decides to cut-in given that the gap with ego car is within its gap acceptance range. This decision means the red vehicle will have a lateral velocity component w.r.t the road. Once the ego vehicle (blue) notices the intention, it will react by adjusting the velocity accordingly. Phase II is \textit{execution}, where the cut-in vehicle actively engages in the lane change, with the ego vehicle accommodating extra space by decelerating where necessary. Finally, Phase III is \textit{completion}, where the cut-in vehicle is fully integrated into the ego vehicle's lane and observes a safe TTC range. 

\section{AV-occupant risk Model}
\label{sec:model}


As introduced in \cite{KOLEKAR2021}, the DRF risk model $R_{DRF}$ is the convolution of the DRF represented by a two-dimensional Gaussian distribution $z(x,y)$ and the driving scene static cost map $C_{S}(x,y)$, given as follows:
\begin{equation}
    R_{DRF} = z(x,y)*C_{S}(x,y).
    \label{eq:risk estimate}
\end{equation}

\begin{figure}[htbp]
\begin{subfigure}{0.24\textwidth}
\includegraphics[width=\linewidth]{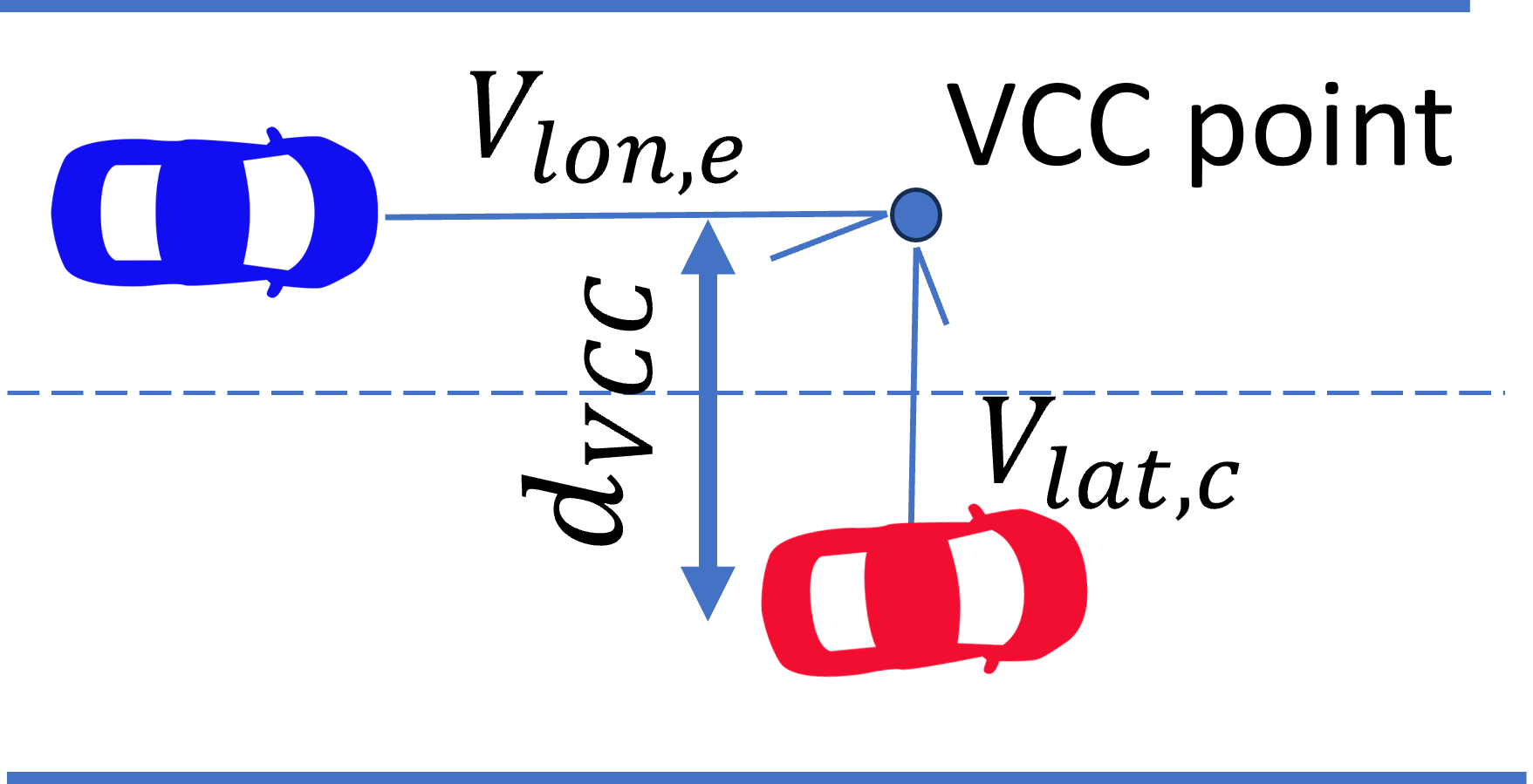} 
\caption{}
\label{fig:model_explanation:vectors}
\end{subfigure}
\hfill 
\begin{subfigure}{0.23\textwidth}
\includegraphics[width=\linewidth]{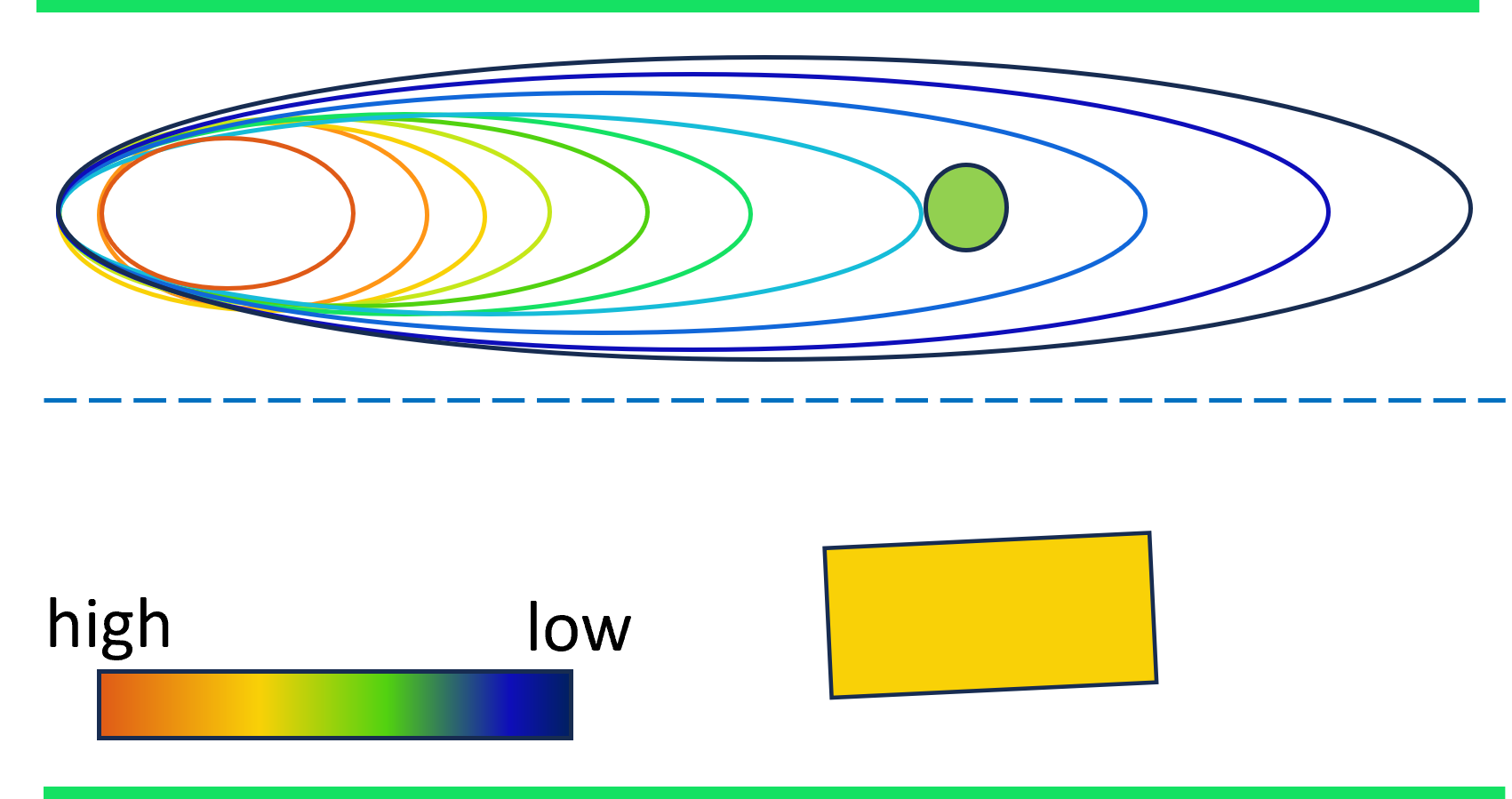}
\caption{}
\label{fig:model_explanation:DRF}
\end{subfigure}
\caption{a) Schematic representation of a cut-in maneuver highlighting the the Virtual Cut-in Collision (VCC) point - marked by the intersection of the lateral velocity vector ($V_{lat,c}$) of the cut-in vehicle and the longitudinal velocity vector ($V_{lon,e}$) of the ego vehicle. It also shows the calculated distance ($d_{VCC}$) between VCC and the cut-in vehicle. 
 b)  Schematic representation of the AVOR costmap with the DRF superimposed on it. The different colors reflect the costs associated with different elements in the scene, as shown by the color bar. The cut-in vehicle is represented as a yellow box, whereas the ego vehicle is situated at the DRF's core. The VCC point is highlighted as a green point within the DRF's field of view, meaning that it would make part of the total risk estimated, contrary to the cut-in vehicle.  }
\label{fig:model_explanation}
\end{figure}

The contribution of our AV-Occupant Risk (AVOR) model $R_{AVOR}$ lies in tackling one of the DRF risk model's main limitations, namely, the inability to account for the unpredictability of dynamic obstacles in the driving scene \cite{Kolekar2020}.  In a speedway, the DRF's field of view extends along the velocity vector of the ego vehicle. A cut-in vehicle in the \textit{neighbouring lane} does not fall within the DRF's field of view as illustrated in Fig. \ref{fig:model_explanation:DRF}, excluding it from the total risk estimated following (\ref{eq:risk estimate}). To account for such critical dynamic information, it is proposed to add a dynamic cost map layer ($C_{D}$) to the existing static cost map ($C_{S}$), represented by
\begin{align}
    R_{AVOR} &= z(x,y) * C_{AVOR}(x,y) \label{eq:risk_estimate_AVOR} \\
    C_{AVOR}(x,y) &= C_{S}(x,y) + C_{D}(x,y) \label{eq:cost_contribution}.
\end{align}

The dynamic cost map would have virtual objects that reflect the potential consequences of dynamic objects in the scene. In the case of cut-in the virtual object is a virtual cut-in collision (VCC) point defined by the intersection of the ego vehicle's longitudinal velocity $V_{lon,e}$ and the cut-in vehicle's lateral velocity $V_{lat,c}$, as shown in Fig. \ref{fig:model_explanation}. The cost at the VCC point is dynamically adjusted \eqref{eq:cost}, increasing as the time-to-arrival (TTA) to the VCC point decreases \eqref{eq:TTA}. This implies that a higher cost is assigned to faster cut-in manoeuvres and, thereby, a greater risk when the VCC point falls within the field of view of the DRF. Hence, the assigned dynamic cost is computed through

\begin{align}
    C_{D}(x_{VCC},\;y_{VCC}) \;\;\; &= \;\;\; \frac{k}{TTA_{VCC}}  \label{eq:cost} \\
    \text{where} \;\;\;\;\; TTA_{VCC} \;\;\; &= \;\;\; \frac{d_{VCC}}{V_{lat,c}}.    \label{eq:TTA}
\end{align}

The baseline cost $k$ is taken to have the same cost as a car from \cite{Kolekar2020}. Parameter values for the DRF and environmental costs are taken from \cite{KOLEKAR2021}.

\begin{figure*}[htbp]
  \centering
  \begin{subfigure}[b]{0.3\textwidth}
    \includegraphics[width=\textwidth]{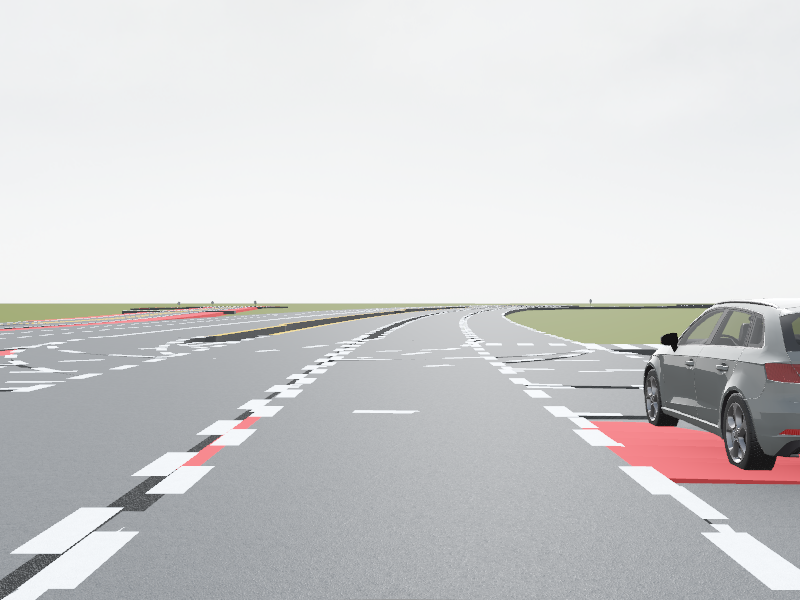}
    \subcaption{\textbf{O}}
    \label{fig:scene_population:O}
  \end{subfigure}
  \hfill 
  \begin{subfigure}[b]{0.3\textwidth}
    \includegraphics[width=\textwidth]{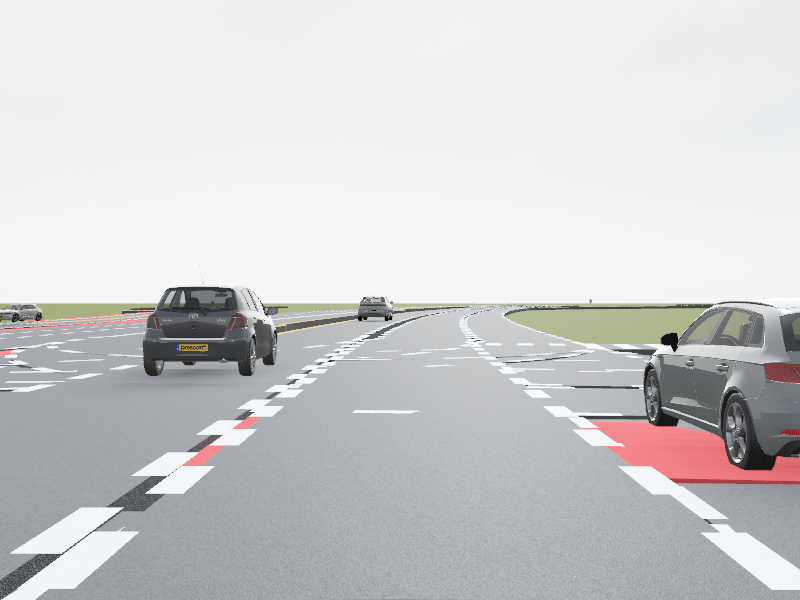}
    \subcaption{\textbf{A}}
    \label{fig:scene_population:A}
  \end{subfigure}
  \hfill 
  \begin{subfigure}[b]{0.3\textwidth}
    \includegraphics[width=\textwidth]{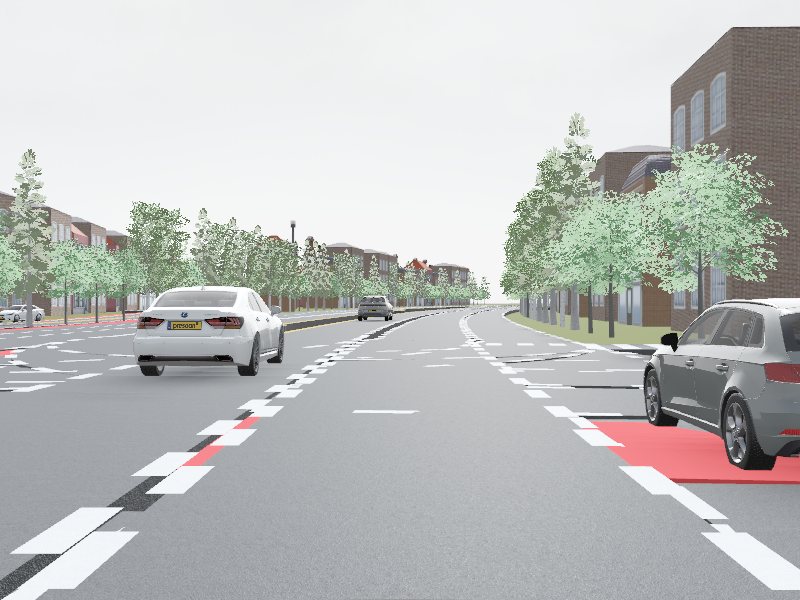}
    \subcaption{\textbf{A+R} }
    \label{fig:scene_population:A+R}
  \end{subfigure}
  \caption{The figure depicts the view from the ego-vehicle as the AV-occupant would see it at the three levels of \textit{scene population}. In Fig. \ref{fig:scene_population:O} the AV-occupant only sees the \textbf{O}bject of interest (level \textbf{O}). In Fig. \ref{fig:scene_population:A} the AV-occupant sees \textbf{A}ll road actors (level \textbf{A}). In Fig. \ref{fig:scene_population:A+R} the AV-occupant sees \textbf{A}ll road actors and \textbf{R}oad furniture, namely buildings and trees (level \textbf{A+R}).}
  \label{fig:scene_population}
\end{figure*}
\section{Experiment}
\label{sec:experiment}

\subsection{Cut-in Scenarios}
\label{sec:scenarios}
\begin{figure}
\centering
\begin{subfigure}{0.24\textwidth}
\includegraphics[width=\linewidth]{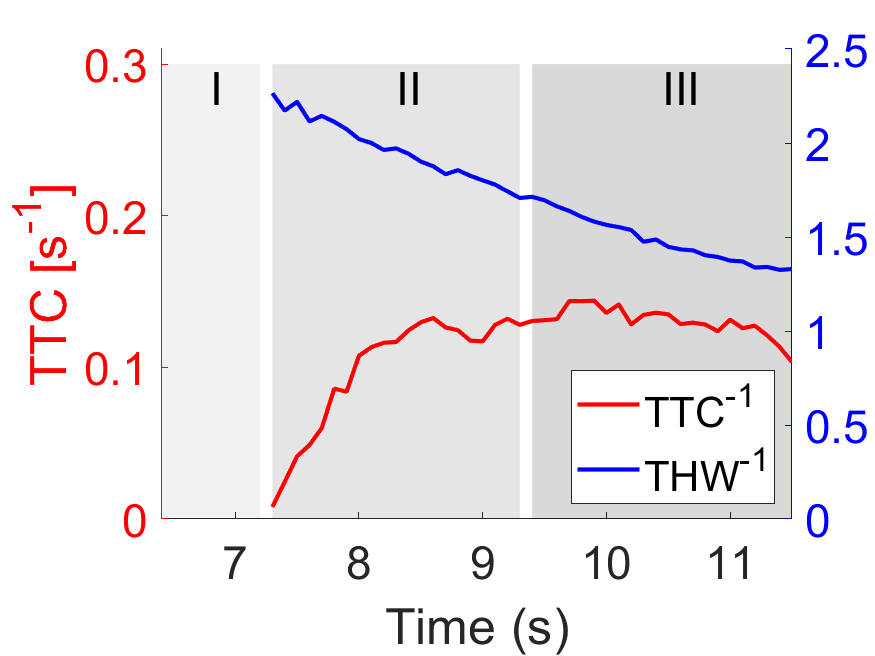} 
\caption{HRS}
\label{fig:SMOS:HRS}
\end{subfigure}
\hfill
\begin{subfigure}{0.24\textwidth}
\includegraphics[width=\linewidth]{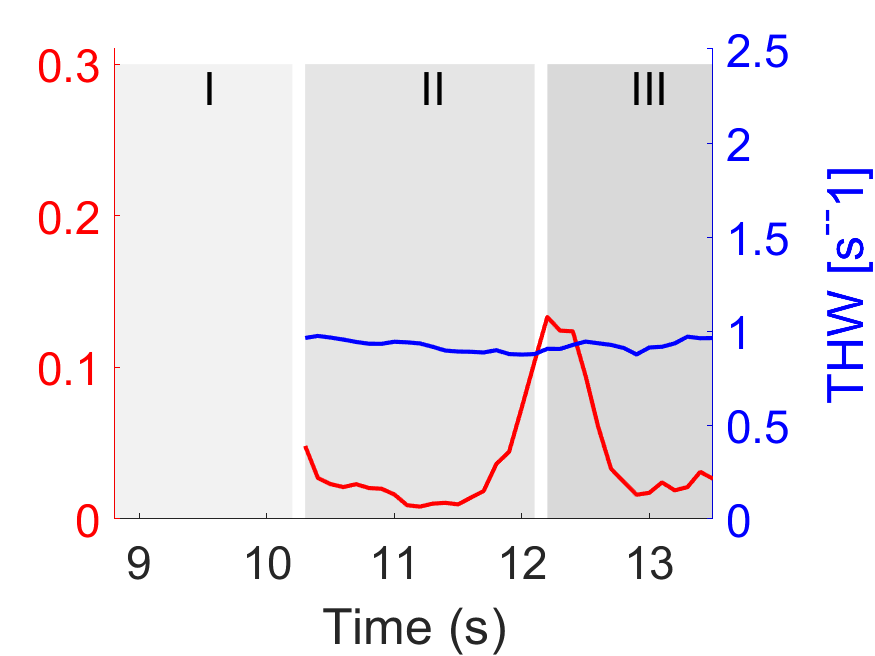}
\caption{LRS}
\label{fig:SMOS:LRS}
\end{subfigure}
\caption{Illustration of the the Surrogate Measures of Safety TTC$^{-1}$ and THW$^{-1}$ for the High Risk Scenario and the Low Risk Scenario respectively. This is given across the different phases of cut-in (I,II,III) as defined in Section \ref{sec:anatomy}. A higher value of both TTC$^{-1}$ and THW$^{-1}$ indicates a higher actual risk.}
\label{fig:SMoS}
\end{figure}

This study uses two (real-life) cut-in scenarios retrieved from the Waymo Open Dataset. These scenarios are labelled as the High-Risk Scenario (HRS) and Low-Risk Scenario (LRS), due to their characteristics and computation of actual risk, illustrated in Fig. \ref{fig:SMoS}. Table \ref{table:cut_in_charachteristics} outlines the characteristics of these cut-in's in numbers. Each scenario takes 20s.

\begin{table}
\centering
\begin{tabular}{c|Ic|c}

 & HRS &  LRS  \\ 
\hline 
\arrayrulecolor{black}\hhline{-|-|-|} 
 Cut-in duration [$s$] & 4.3 & 4.9 \\
\hline
Average $V_{lat}$ [$ms^{-1}$] & 0.757 & 0.3049  \\
\hline
$V_{lat_{max}}$ [$ms^{-1}$] & 1.275 & 0.7419  \\
\hline
 Average $a_{lat}$ [$ms^{-2}$]  & 0.015 & 0.0086  \\
\hline
Initial cut-in distance [m] & 8 & 12.8 \\
\hline
\end{tabular}
\caption{Kinematic characteristics of the two tested cut-in scenarios: the High-Risk Scenario (HRS) and the Low-Risk Scenario (LRS).}
\label{table:cut_in_charachteristics}
\end{table}

\subsection{Partipants}
There were 18 participants, with an average age of 27 years and a standard deviation of 5 years.  All participants exhibit a driver's license with an average of 8 years of driving experience.

\subsection{Setup}
\begin{figure}[htbp]
\centering
\includegraphics[width=0.4\textwidth]{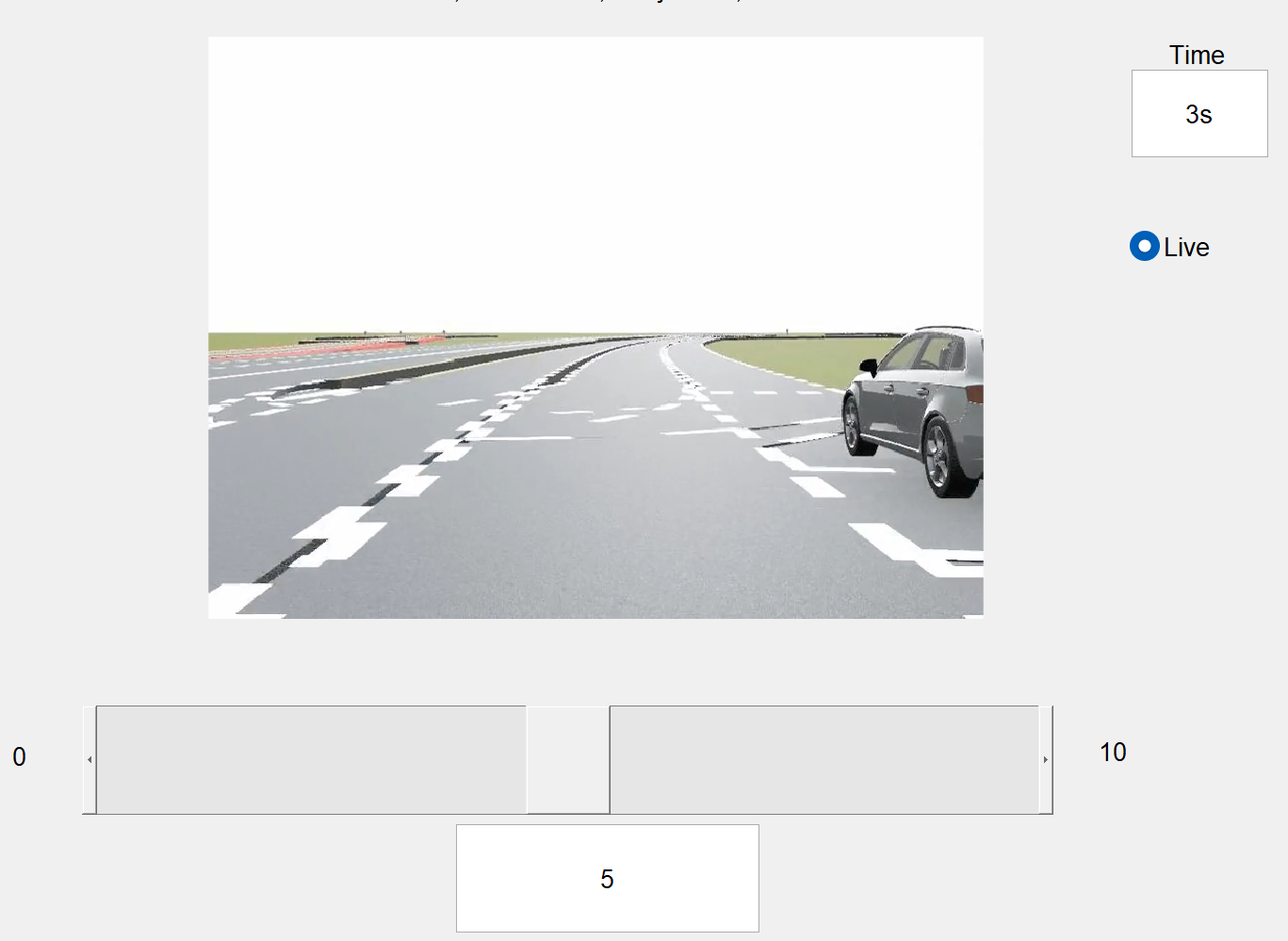}
\caption{The GUI used in the user experiment}
\label{fig:cGUI}
\end{figure}

The Waymo scenario's were visualised using Simcenter Prescan, using an in-house developed Real2Sim pipeline. The scenarios were played inside a Matlab GUI illustrated in Fig. \ref{fig:cGUI} with a slider bar that would record the user's Subjective Risk Rating (SRR) in real-time. A live button would be on when the video was streaming live and subjective rating data was recorded. Time feedback was presented through a  timer. Slider value feedback was given as a number appearing below the slider bar.

\subsection{Instructions}
Subjects were asked to read the briefing and sign a consent form on arrival. They were instructed to give their SRR, assuming they were an AV occupant riding through the presented scenarios. They were informed that a rating of 5 is neutral, 10 is unacceptable, and 0 is non-existent risk.    


\subsection{Independent Variables}

There are two independent variables tested. 
\begin{enumerate}
    \item Scenario risk
    \item Scene population
\end{enumerate}

The first is scenario-risk, with two levels of risk considered as explained in Section \ref{sec:scenarios}.  The second is the scene-population, with three different levels: only Objects of interest (O), All road actors (A) and All road actors with Road furniture (A+R), as illustrated in Fig. \ref{fig:scene_population}.

\subsection{Dependent variables}

\subsubsection{Statistical Analysis of Subjective Risk Ratings}
For quantifying and characterising the measured differences in SRRs, the \textit{average SRR} across the three phases of cut-in as described in Fig. \ref{fig:cut-in_anatomy} is analysed.

A two-way repeated-measures Analysis of Variance (ANOVA) was performed to evaluate the statistical significance of differences in these metrics. There are two factors in the design of this experiment: Scenario and Scene population. Assumptions for this statistical test to be valid include 1) \textit{normality}, which was tested using the Shapiro-Wilk test and 2)  \textit{sphericity}, which was corrected for using the Greenhouse-Geisser correction in case of a violation.

\begin{figure*}
\centering
\begin{subfigure}{0.3\textwidth}
\includegraphics[width=\linewidth]{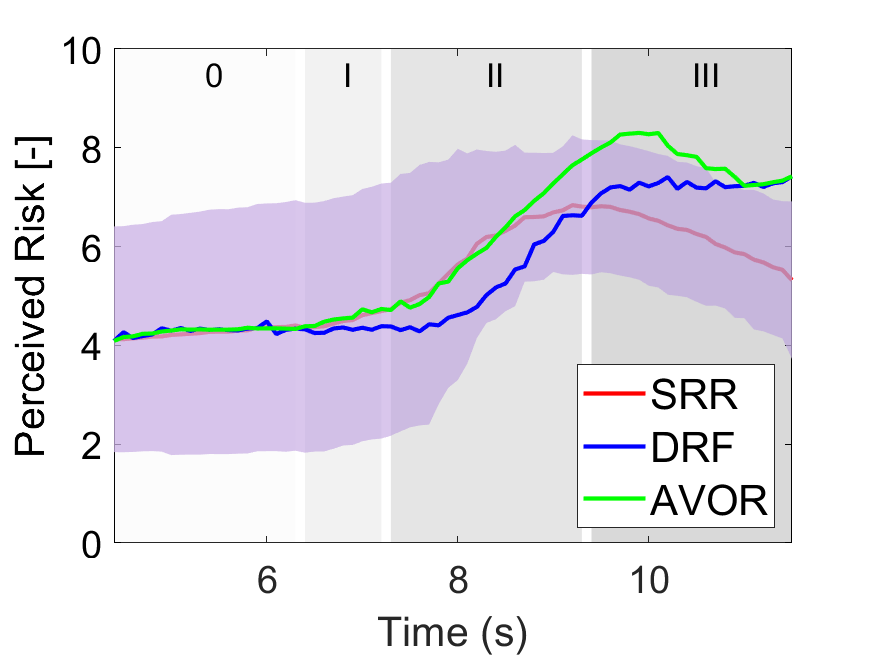} 
\caption{HRS, O}
\label{fig:Model_evaluation:HRS_O}
\end{subfigure}
\hfill
\begin{subfigure}{0.3\textwidth}
\includegraphics[width=\linewidth]{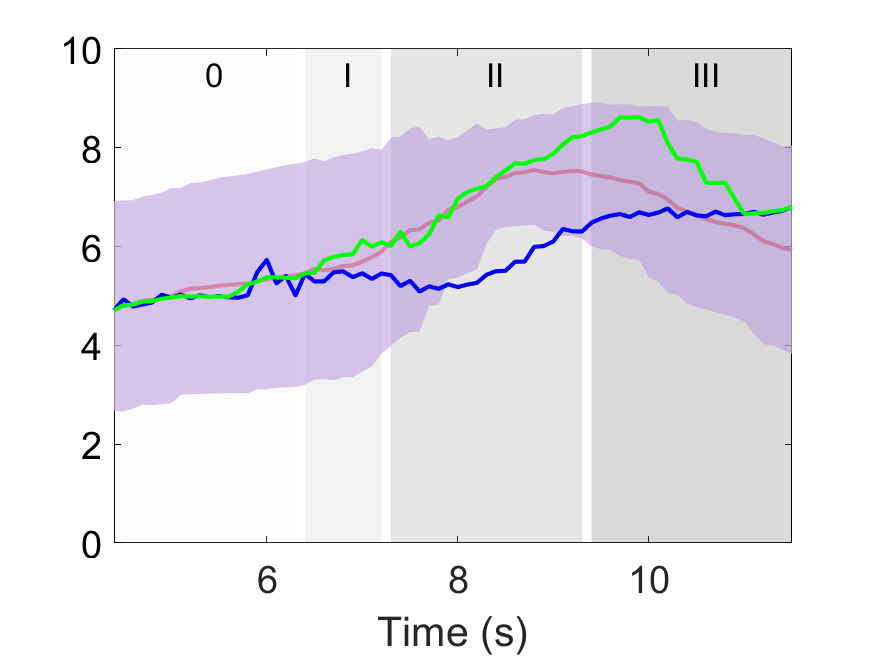}
\caption{HRS, A}
\label{fig:Model_evaluation_HRS_A}
\end{subfigure}
\hfill
\begin{subfigure}{0.3\textwidth}
\includegraphics[width=\linewidth]{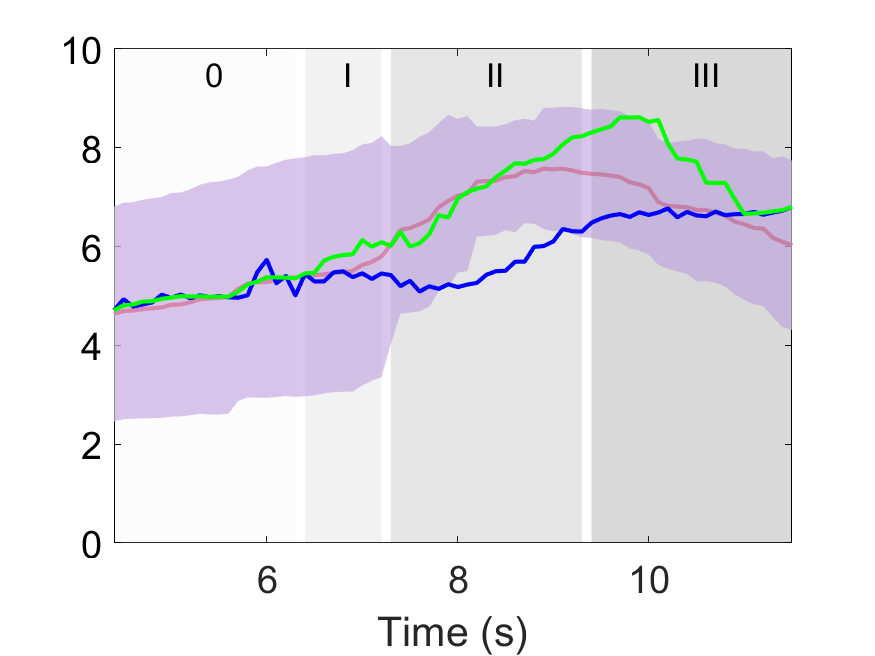} 
\caption{HRS, A+R}
\label{fig:Model_evaluation_HRS_A+R}
\end{subfigure}
\begin{subfigure}{0.3\textwidth}
\includegraphics[width=\linewidth]{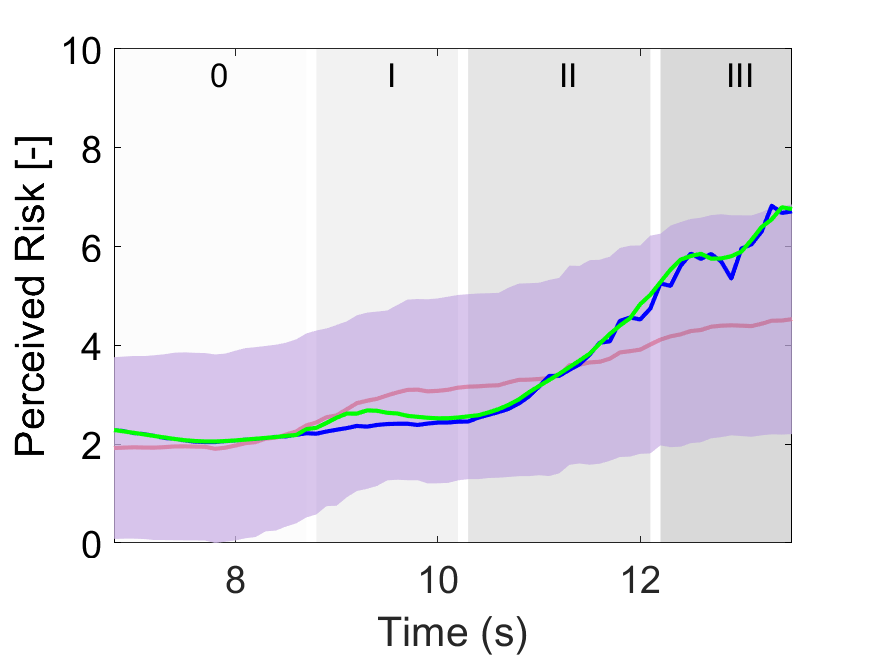} 
\caption{LRS, O}
\label{fig:Model_evaluation_LRS_O}
\end{subfigure}
\hfill
\begin{subfigure}{0.3\textwidth}
\includegraphics[width=\linewidth]{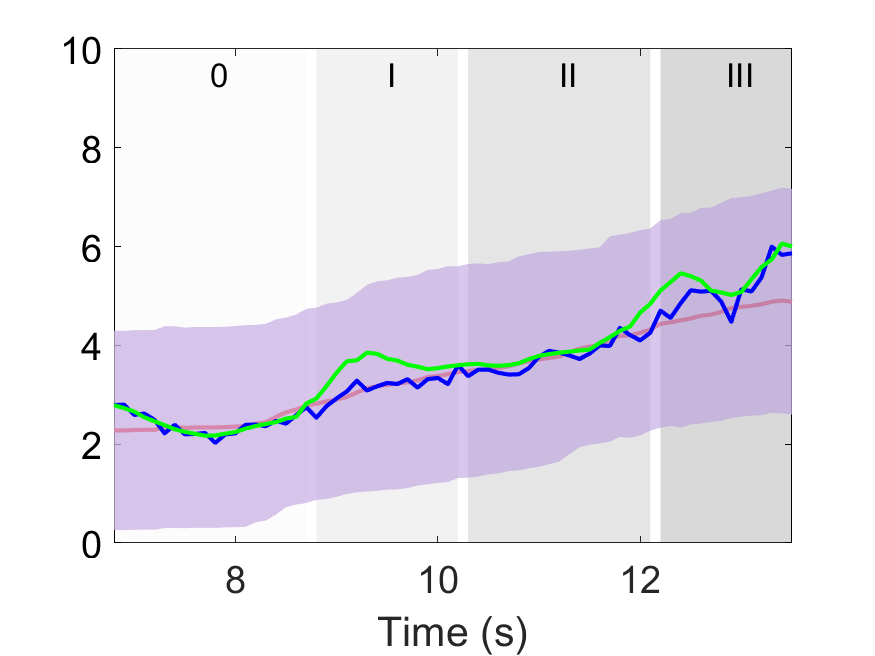}
\caption{LRS, A}
\label{fig:Model_evaluation_LRS_A}
\end{subfigure}
\hfill
\begin{subfigure}{0.3\textwidth}
\includegraphics[width=\linewidth]{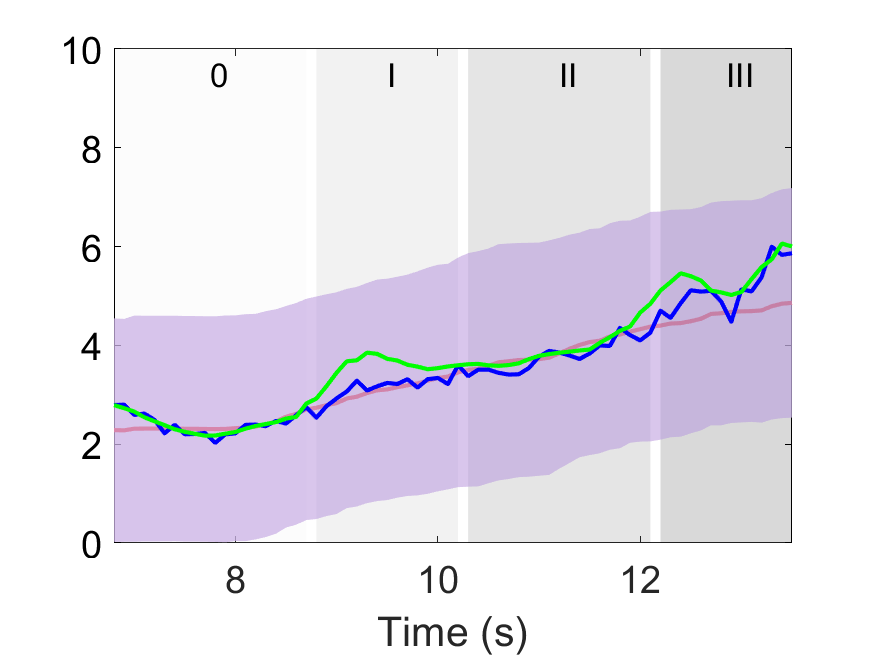} 
\caption{LRS, A+R}
\label{fig:Model_evaluation_A+R}
\end{subfigure}
\caption{Illustration of the six evaluated conditions: High-Risk Scenarios (HRS) and Low-Risk Scenarios (LRS) across scene populations O, A, and A+R. a), b) and c) show HRS conditions, while d) e) and f) LRS conditions. Each plot displays the average and standard deviation of Subjective Risk Ratings (SRR) in red. The normalized and scaled risk estimates: the DRF risk estimate is in blue, and the AVOR risk estimate is in green. Grey boxes outline the specific cut-in phases as detailed in Section \ref{sec:anatomy}.}
\label{fig:Model_evaluation}
\end{figure*}

\subsubsection{AVOR model evaluation}
\label{sec:AVOR model validation}

The unit-less modelled risk values were normalised to the maximum risk value per condition to compare the risk trend, employing min-max normalization \eqref{eq:min-max-normalisation}. This technique adjusts the ranges of modelled perceived risk by scaling them to a [0,10] range, consistent with subjective risk ratings.

\begin{equation}
    \hat{z}_{k} = \frac{z_{k} - z_{min}}{z_{max}-z_{min}} +\bar{c}_{SRR}
    \label{eq:min-max-normalisation}
\end{equation}

Here $\hat{z}_{k}$ is the scaled perceived risk model output; $z_{max}$ and $z_{min}$ are the maximum and minimum values of the modelled perceived risk data in a given condition. The constant $\bar{c}_{SRR}$ is added, quantifying the average perceived risk 2 seconds before the cut-in manoeuvre (Phase 0).

To evaluate the performance of the models, the Root Mean Square Error (RMSE) is computed between the estimated outputs, $\hat{z}_{k}$, and the average SRR for each condition across the defined cut-in phases (see Fig. \ref{fig:cut-in_anatomy}). This approach aims to capture the general trend in SRR, prioritizing aggregate behaviour over individual differences. 

\subsection{Hypothesis}
There are four hypotheses formulated for this experiment.

\begin{description}
    \item[\textbf{H.I}] The average SRR will be larger for the HRS than for the LRS.
    \item[\textbf{H.II}] The average SRR will be higher for the  A than the O scene population condition.
    \item[\textbf{H.III}] The average SRR will be unaffected by road furniture, i.e. no difference between A and A+R conditions.
    \item[\textbf{H.IV}] All subjects' perceived risk increases at phase I, demonstrated by  $\Delta$ SRR $>$ 0.5 rating points.
\end{description}





\section{Results}
\label{sec:results}

The results section is structured into two segments: first, we conduct a statistical analysis of Subjective Risk Ratings (SRR) per cut-in phase to examine the impact of scenario types (HRS and LRS) and scene populations (O, A, A+R) on perceived risk. Second, we delve into the performance assessment of the AVOR model across these variations.

\subsection{Statistical Analysis of Risk Ratings}

Statistically significant effects of scenario type on average SRR are observed across all three phases. The continuous values of SRR can be seen in red in Fig. \ref{fig:Model_evaluation}. In Phase I, HRS are associated with a 1.7-point increase in average SRRs compared to low-risk scenarios (LRS), confirmed by F(2, 12) = 20.6, p $<$ 0.01. Phase II continues this trend, with HRS leading to a 2.7-point rise in average SRRs, as shown by F(2,12) = 110, p$<$0.01. Similarly, Phase III reports a 2.1-point augmentation in average SRRs due to HRS, supported by F(2,12) = 52, p$<$0.01.  without violating normality assumptions.

The influence of scene population on average SRRs is significantly noted in Phase II. The inclusion of additional road actors (scenario A) and road furniture (scenario A+R) results in average SRRs being elevated by 0.6 and 1.1 points, respectively, above the baseline scenario (O). This effect is statistically significant, as evidenced by F(2,12) = 3.52, p$<$0.05, with pairwise comparisons revealing notable differences between scenarios O and A (p$<$0.05) and O and A+R (p$<$0.01). These results highlight the role of scene complexity, in addition to scenario type, in modulating perceived risk.

\subsection{AVOR model evaluation}

The comparison between the AVOR and DRF models, along with SRR, is depicted in Figure \ref{fig:Model_evaluation}, with corresponding RMSE values detailed in Table \ref{table:RMSE_results}.

In scene population O during Phase I, the AVOR model outperforms the DRF model for both HRS and LRS, showing RMSE improvements of 0.03 and 0.14, respectively. This enhancement reflects the AVOR model's responsiveness to the virtual cut-in collision point, effectively capturing the increase in perceived risk at the beginning of Phase I. For HRS in Phase II, the AVOR model continues to provide a more accurate risk assessment by timely adjusting to perceived risk increases, unlike the DRF model, which exhibits a response lag. By Phase III, however, neither model effectively tracks the perceived risk trend.

For scene population A in HRS during Phase I, the AVOR model precisely captures the uptrend in perceived risk at the cut-in's commencement, unlike the DRF model, which fails to accurately reflect this rise. Phase II highlights the AVOR model's significant advancement, showing a 54\% improvement in perceived risk estimation over the DRF model, which shows a perceived risk reduction correlating with the ego vehicle's deceleration, as evidenced by the THW in Figure \ref{fig:SMOS:HRS}. In LRS during Phase I, the AVOR model's risk estimates exceed actual perceived risk, while the DRF model appears to align with the actual risk trend across all phases more closely.

Regarding the A+R scene population, the model outputs remain consistent with those for scene A, given that the added elements—buildings and trees—are beyond the risk field's visual range. However, minor variations in SRR are noted, impacting RMSE calculations. 

\begin{table*}[ht]
\centering
\captionsetup{justification=centering}
\caption{This table presents the Root Mean Square Error (RMSE) values for perceived risk estimates from the DRF and AVOR models. The computation spans scene populations (O, A, A+R), scenarios (HRS, LRS), and cut-in phases (I,II,III).}
\begin{tabular}{ccccccccccccccccccc}
\toprule
 & \multicolumn{6}{c}{O} & \multicolumn{6}{c}{A} & \multicolumn{6}{c}{A+R} \\ 
\cmidrule(lr){2-7} \cmidrule(lr){8-13} \cmidrule(lr){2-7} \cmidrule(lr){14-19}
 & \multicolumn{3}{c}{HRS} & \multicolumn{3}{c}{LRS} & \multicolumn{3}{c}{HRS} & \multicolumn{3}{c}{LRS}
 & \multicolumn{3}{c}{HRS} & \multicolumn{3}{c}{LRS}
 \\
\midrule
 & I & II & III & I & II & III & I & II & III & I & II & III & I & II & III & I & II & III \\ \hline
 DRF & 0.21 & 0.79 & 1.14 & 0.55 & 0.47 &1.62 & 0.43 & 1.64 &0.56 & 0.13 & 0.12 & 0.83 & 0.36 & 1.66 & 0.55 & 0.13 & 0.12 & 0.98 \\ \hline
AVOR & 0.18 & 0.46 & 1.59  & 0.41 & 0.47
& 1.71 & 0.32 & 0.58 & 1.08 & 0.55 & 0.14 & 0.57 & 0.41 & 0.57 & 0.98 & 0.43 & 0.13 & 0.54 \\
\bottomrule
\end{tabular}
\label{table:RMSE_results}
\end{table*}

\section{Discussion}
\label{sec:discussion}

The average perceived risk aligns with the average actual risk of an entire event. Actual risk metrics such as THW and TTC (see Table \ref{table:cut_in_charachteristics} \& Fig. \ref{fig:SMoS}) were used to select the scenarios from Waymo, resulting in HRS and LRS. These surrogate measures of safety have successfully estimated the perceived risk of entire events or driving scenarios in literature \cite{HE2022,Kondoh2008}. Specifically, THW significantly impacts stress levels, while the inverse of Time-To-Collision (TTC$^{-1}$) often prompts braking actions by following drivers \cite{Kondoh2008}. The results presented in this paper are aligned, with the HRS scenario consistently eliciting a significantly higher perceived risk across all cut-in phases, with an average of 2.1 rating points increase, validating \textbf{H.I}. However, surrogate safety measures do not capture the \textit{continuous} aspect of perceived risk. The average correlation between the inverse of Time Headway (THW$^{-1}$), Time-To-Collision (TTC$^{-1}$), and SRR stands at 0.4, suggesting instances where risk may be either overestimated or underestimated \cite{CHARLTON201450}. 



 It is no surprise, increasing road density would increase perceived risk, reflected by a significant increase between O and A for phase II and validating \textbf{H.II}. This trend is particularly pronounced in High-Risk Scenarios (HRS) but is inadequately captured by the DRF model. Moreover, the addition of road furniture, such as trees, results in an average perceived risk increase of 1.1 rating points from O compared to only 0.6 rating points from O to A. This is a phenomenon that neither the AVOR nor the DRF models effectively recognize due to their inability to consider objects outside their visual field. Nevertheless, the transition from A to A+R does not yield a statistically significant difference, leading to the acceptance of \textbf{H.III}. The presence of buildings and trees introduces an element of optical flow; while some participants reported an increase in perceived risk due to this factor, others felt it reduced their risk perception, associating it with lower speed limits and safer driving conditions.


For the collected experiment data, it is found that 76\% of the SRRs collected show a significant risk perception increase (defined as a minimum increase of 0.5 ratings) during phase I, as illustrated in Fig. \ref{fig:Model_evaluation} (validating \textbf{H.IV}).  However, the DRF model fails to recognize this uptick in risk perception in the absence of other road actors, in contrast to the AVOR model, which accurately estimates this critical increase.  In the context of driving, the timely identification of perceived risk is essential. It could establish a critical threshold between a comfortable ride and an AV occupant losing trust and rejecting the AV altogether.



\section{Conclusion}
\label{sec:conclusion}

The aim of this study is to accurately model perceived risk during cut-in manoeuvres. A novel AV-Occupant Risk (AVOR) model is introduced that enhances the Driver's Risk Field (DRF) model by incorporating the uncertainties of dynamic objects in a novel dynamic cost map. Specifically for cut-in manoeuvres, a virtual cut-in collision point is introduced. To evaluate our model a user experiment was performed with 18 subjects across two realistic Waymo cut-in scenarios—low risk and high risk—spanning three distinct road population levels. A majority of the subjective risk responses (76\%) indicate an increase in perceived risk at the onset of cut-in manoeuvres. However, this trend is not detected by the DRF. Conversely, the AVOR model demonstrated a significant improvement in matching perceived risk during the early stages of cut-ins, enhancing modelling accuracy by up to 54\%. This is a promising result and further evaluation studies are necessary to assess the model's performance across a wider array of realistic driving scenarios.

\section*{Acknowledgment}
This project has received funding from the Flemish Agency for Innovation and Entrepreneurship (VLAIO) under research project No. HBC.2021.0939 (BECAREFUL - BELGIAN CONSORTIUM FOR ENHANCED SAFETY \&
COMFORT PERCEPTION OF FUTURE AUTONOMOUS
VEHICLES). We want to express our gratitude to Jean-Pierre Allamaa for his insightful comments that contributed to improving the quality of this paper.



%

\bibliographystyle{plain}
\bibliography{ref}

\begin{thebibliography}{10}

\bibitem{1994NHTSA}
Lane change/merge crashes: Problem size assessment and statistical description.
\newblock {\em U.S. Department of Transportation National Highway Traffic Safety Administration}, 12, 1994.

\bibitem{SOTIF}
Challenges in autonomous vehicle testing and validation.
\newblock {\em Safety of the Intended Functionality}, pages 125--142, 2020.

\bibitem{ISO2631}
ISO 2631.
\newblock Mechanical vibration and shock – evaluation of human exposure to whole-body vibration, standard.
\newblock {\em International Organization for Standardization, Geneva, CH}, 2001.

\bibitem{CHARLTON201450}
Samuel~G. Charlton, Nicola~J. Starkey, John~A. Perrone, and Robert~B. Isler.
\newblock What’s the risk? a comparison of actual and perceived driving risk.
\newblock {\em Transportation Research Part F: Traffic Psychology and Behaviour}, 25:50--64, 2014.

\bibitem{deWinkel2022}
Ksander de~Winkel, Tugrul Irmak, Riender Happee, and Barys Shyrokau.
\newblock Standards for passenger comfort in automated vehicles: Acceleration and jerk.
\newblock {\em Applied Ergonomics}, 106, 09 2022.

\bibitem{DEWINTER2023235}
Joost {de Winter}, Jim Hoogmoed, Jork Stapel, Dimitra Dodou, and Pavlo Bazilinskyy.
\newblock Predicting perceived risk of traffic scenes using computer vision.
\newblock {\em Transportation Research Part F: Traffic Psychology and Behaviour}, 93:235--247, 2023.

\bibitem{DelRe2022}
Enrico del Re and Cristina Olaverri-Monreal.
\newblock Implementation of road safety perception in autonomous vehicles in a lane change scenario.
\newblock In {\em 2022 IEEE International Conference on Vehicular Electronics and Safety (ICVES)}, pages 1--6, 2022.

\bibitem{ERTRAC2019}
ERTRAC~Working Group.
\newblock Connected automated driving roadmap.
\newblock {\em Connectivity and Automated Driving}, 2019.

\bibitem{GUO2021240}
Yuxi Guo, Qinyu Sun, Yanqi Su, Yingshi Guo, and Chang Wang.
\newblock Can driving condition prompt systems improve passenger comfort of intelligent vehicles? a driving simulator study.
\newblock {\em Transportation Research Part F: Traffic Psychology and Behaviour}, 81:240--250, 2021.

\bibitem{HE2022}
Xiaolin He, Jork Stapel, Meng Wang, and Riender Happee.
\newblock Modelling perceived risk and trust in driving automation reacting to merging and braking vehicles.
\newblock {\em Transportation Research Part F: Traffic Psychology and Behaviour}, 86:178--195, 2022.

\bibitem{Irmak2022}
Tugrul Irmak, Ksander de~Winkel, Daan Pool, Heinrich~H. Bulthoff, and Riender Happee.
\newblock Individual motion perception parameters and motion sickness frequency sensitivity in fore-aft motion.
\newblock {\em Experimental Brain Research}, 6:1727--1745, 9 2021.

\bibitem{iso26262}
ISO.
\newblock 26262: 2018: Road vehicles—functional safety.
\newblock {\em British Standards Institute}, 12, 2018.

\bibitem{Baekgyu2017}
Baekgyu Kim, Yusuke Kashiba, Siyuan Dai, and Shinichi Shiraishi.
\newblock Testing autonomous vehicle software in the virtual prototyping environment.
\newblock {\em IEEE Embedded Systems Letters}, 9(1):5--8, 2017.

\bibitem{Kolekar2020}
Sarvesh Kolekar, Joost {de Winter}, and David Abbink.
\newblock Human-like driving behaviour emerges from a risk-based driver model.
\newblock {\em Nature Communications}, 11, 2020.

\bibitem{KOLEKAR2021}
Sarvesh Kolekar, Bastiaan Petermeijer, Erwin Boer, Joost {de Winter}, and David Abbink.
\newblock A risk field-based metric correlates with driver’s perceived risk in manual and automated driving: A test-track study.
\newblock {\em Transportation Research Part C: Emerging Technologies}, 133:103428, 2021.

\bibitem{Kondoh2008}
Takayuki Kondoh, Tomohiro Yamamura, Satoshi Kitazaki, Nobuyuki Kuge, and Erwin Boer.
\newblock Identification of visual cues and quantification of drivers' perception of proximity risk to the lead vehicle in car-following situations.
\newblock {\em Journal of Mechanical Systems for Transportation and Logistics}, 1:170--180, 04 2008.

\bibitem{Li2021}
Yang Li, Keqiang Li, Yang Zheng, Bernhard Morys, Shuyue Pan, and Jianqiang Wang.
\newblock Threat assessment techniques in intelligent vehicles: A comparative survey.
\newblock {\em IEEE Intelligent Transportation Systems Magazine}, 13(4):71--91, 2021.

\bibitem{Mohajer2020}
Navid Mohajer, Saeid Nahavandi, Hamid Abdi, and Zoran Najdovski.
\newblock Enhancing passenger comfort in autonomous vehicles through vehicle handling analysis and optimization.
\newblock {\em IEEE Intelligent Transportation Systems Magazine}, 13(3):156--173, 2021.

\bibitem{MONTORO2019}
Luis Montoro, Sergio~A. Useche, Francisco Alonso, Ignacio Lijarcio, Patricia Bosó-Seguí, and Ana Martí-Belda.
\newblock Perceived safety and attributed value as predictors of the intention to use autonomous vehicles: A national study with spanish drivers.
\newblock {\em Safety Science}, 120:865--876, 2019.

\bibitem{MORAN2019309}
Caroline Moran, Joanne~M. Bennett, and Prasannah Prabhakharan.
\newblock Road user hazard perception tests: A systematic review of current methodologies.
\newblock {\em Accident Analysis \& Prevention}, 129:309--333, 2019.

\bibitem{MULLAKKALBABU2020102716}
Freddy~A. Mullakkal-Babu, Meng Wang, Xiaolin He, Bart {van Arem}, and Riender Happee.
\newblock Probabilistic field approach for motorway driving risk assessment.
\newblock {\em Transportation Research Part C: Emerging Technologies}, 118:102716, 2020.

\bibitem{PRASETIO2023}
Eko~Agus Prasetio and Cintia Nurliyana.
\newblock Evaluating perceived safety of autonomous vehicle: The influence of privacy and cybersecurity to cognitive and emotional safety.
\newblock {\em IATSS Research}, 47(2):160--170, 2023.

\bibitem{Tariq2023}
Faizan~M. Tariq, David Isele, John~S. Baras, and Sangjae Bae.
\newblock Rcms: Risk-aware crash mitigation system for autonomous vehicles.
\newblock In {\em 2023 IEEE 26th International Conference on Intelligent Transportation Systems (ITSC)}, 2023.

\bibitem{Xu2018}
Zhigang Xu, Kaifan Zhang, Haigen Min, Zhen Wang, Xiangmo Zhao, and Peng Liu.
\newblock What drives people to accept automated vehicles? findings from a field experiment.
\newblock {\em Transportation Research Part C: Emerging Technologies}, 95:320--334, 2018.

\end{thebibliography}

\end{document}